\documentclass[twoside]{article}

\usepackage{PRIMEarxiv}
\usepackage{natbib}
\usepackage[utf8]{inputenc} 
\usepackage[T1]{fontenc}    
\usepackage{geometry}
\usepackage[dvipsnames]{xcolor}
\usepackage{algorithm}
\usepackage{algorithmic}
\usepackage{amsmath}
\usepackage{tabularx}
\usepackage{comment}
\usepackage{longtable}
\usepackage{multirow}
\usepackage{comment}
\usepackage{hyperref}       
\usepackage{url}            
\usepackage{booktabs}       
\usepackage{amsfonts}       
\usepackage{nicefrac}       
\usepackage{microtype}      
\usepackage{fancyhdr}       
\usepackage{graphicx}       
\usepackage{colortbl}
\usepackage{listings}
\usepackage[toc,page]{appendix}

\title{Information-Consistent Language Model Recommendations through Group Relative Policy Optimization
}

\author{
  Sonal Prabhune, Balaji Padmanabhan, and Kaushik Dutta  \\
  \texttt{\{saprabhune@usf.edu, bpadmana@umd.edu, duttak@usf.edu\}} \\
}

\begin{document}
\maketitle

\begin{abstract}
Large Language Models (LLMs) are increasingly deployed in business-critical domains such as finance, education, healthcare, and customer support, where users expect consistent and reliable recommendations. Yet LLMs often exhibit variability when prompts are phrased with minor differences, even when semantically equivalent. Such inconsistency undermines trust, complicates compliance, and disrupts user experience. While personalization is desirable in certain contexts, many enterprise scenarios, such as HR onboarding, customer support, or policy disclosure, require invariant information delivery regardless of phrasing or prior conversational history. Existing approaches, including retrieval-augmented generation (RAG) and temperature tuning, improve factuality or reduce stochasticity, but cannot guarantee stability across equivalent prompts. In this paper, we propose a reinforcement learning framework based on Group Relative Policy Optimization (GRPO) to directly optimize for consistency. Unlike prior applications of GRPO, which have been limited to reasoning and code generation, we adapt GRPO to enforce the stability of information content across groups of semantically equivalent prompts. We introduce entropy-based helpfulness and stability rewards, treating prompt variants as groups and resetting conversational context to isolate phrasing effects. Experiments on investment and job recommendation tasks show that our GRPO-fine-tuned model reduces variability compared to the baseline LLM model. To our knowledge, this is a novel application of GRPO for aligning LLMs toward information consistency, reframing variability not as an acceptable feature of generative diversity, but as a correctable flaw in enterprise deployments.
\end{abstract}

\section{INTRODUCTION}

Large Language Models (LLMs) such as Llama-3 are increasingly deployed in domains requiring decision support and recommendations. A critical requirement in these deployments is that the AI provides \textbf{consistent and reliable outputs} regardless of how a user phrases a prompt. Consistency is not merely a technical consideration—it underpins trust, usability, and compliance in business applications. In practice, organizations depend on stable outputs to ensure operational reliability, regulatory adherence, brand integrity, and user satisfaction. At the same time, consistent behavior is also essential to safeguard fairness and prevent systemic harms. 

It is important to acknowledge that consistency is not always universally desirable. In some settings, such as personalized learning platforms or health coaching, users benefit when the system tailors its responses to their unique profiles and histories. Here, variation in outputs can reflect meaningful personalization rather than unreliability. However, there are many business-critical scenarios where consistency must prevail regardless of how well the LLM knows the user or their prior interaction context. For example, the use of LLMs for answering organizational policy-related questions, personal financial planning questions, and educational planning questions are common applications. In such applications, providing a consistent answer is important in building the user trust on LLM based answers. In these cases, personalization should not alter the essential information content of the response. For example, in human resources onboarding, new employees should always receive the same explanation of company policies; in customer support, answers about product warranty coverage should remain unchanged no matter how the question is phrased; and in compliance-driven settings, financial disclosures or insurance terms must be delivered identically to every user. These are cases where personalization may add conversational nuance but should never compromise the consistency of core information. 

Solutions such as RAG \cite{lewis2020rag} have been proposed as a way to support consistency by grounding answers in external knowledge. While effective in many contexts, RAG does not fully eliminate inherent inconsistencies in LLM behavior. For instance, consider two applicants preparing for the same job interview and querying an LLM-powered assistant: one asks, ``What are the most common data structures I should review?'' while another asks, ``Which data structures should I prepare for in coding interviews?'' Even when both queries are backed by the same retrieved documents, the LLM may generate divergent lists or emphasize different topics. Such inconsistencies create confusion and reduce trust in the system, especially in high-stakes contexts like interview preparation. Literature has often embraced this variability as an acceptable property of generative models, framing it as diversity in responses \cite{holtzman2019curious} and proposing workarounds such as RAG or temperature tuning to control stochasticity. However, this acceptance does not resolve the fundamental issue. Simply lowering temperature or relying on retrieval does not guarantee that semantically equivalent prompts will produce consistent outputs. For many business and educational applications, overlooking this inconsistency is untenable. 

Across industries such as finance, education, healthcare, and customer services, unpredictable or inconsistent responses can have significant consequences and may not help build the required user trust in using such systems. A bank providing different disclosures depending on phrasing, risks compliance failures; a chatbot that answers the same question differently for two customers undermines confidence in customer support; and in educational or hiring contexts, demographic attributes such as gender introducing unintended inconsistencies result in ethical issues. In particular, job recommendation scenarios highlight both the operational and ethical risks of inconsistency and demonstrate the need for robust methods to enforce stable and equitable outputs.

While contextual methods such as RAG can help improve consistency by grounding model responses in relevant documentation, their applicability depends on whether contextual retrieval is available at query time. In enterprise deployments, RAG can reduce hallucinations and tie answers to authoritative sources, thereby increasing factual consistency. However, many user interactions occur without such context—when individuals query general-purpose assistants directly, without attached documents or retrieval layers. In such settings, the model must still produce internally consistent responses across semantically equivalent prompts, regardless of who is asking the question. In this paper, we focus on this latter class of scenarios—direct, context-free user interactions—and leave the extension to contextual or retrieval-grounded querying as a direction for future work.



In addition to operational and reputational incentives, \textbf{legal risk} increasingly drives the demand for consistency in AI systems. Recent lawsuits highlight how inconsistent model behavior can expose organizations to liability. In \emph{Mobley v.\ Workday} ~\cite{hklaw2025workdaybias} and \emph{Harper v.\ Sirius XM} ~\cite{fisherphillips2024sirius}, allegations of hiring bias highlight how inconsistent or opaque decision patterns can be interpreted as unlawful disparate treatment. Cases involving chatbot misinformation, such as the British Columbia tribunal’s finding of liability for an AI-generated misstatement ~\cite{americanbar2024chatbotliability}, further illustrate that even occasional unreliability can create legal exposure. Likewise, the wrongful-death claim in \emph{Raine v.\ OpenAI} ~\cite{raine2024openai} points to the risks posed by inconsistent adherence to safety-critical behaviors. These examples demonstrate that inconsistency undermines fairness, defensibility, and regulatory compliance, turning consistency from a technical preference into a \textbf{legal imperative}.

It is important to note that in certain cases, \textbf{context can assist consistency}. RAG frameworks, for example, can anchor model responses to verified documentation or structured data, reducing factual drift and stabilizing answers across paraphrased prompts. When properly implemented, RAG can effectively “ground” responses in organizational context—such as product manuals, policy databases, or compliance archives—ensuring that the same underlying information supports each response.
However, not all scenarios can rely on contextual retrieval. In many real-world interactions, users directly query an LLM without supplementary context, expecting consistent and accurate answers purely from the model’s internal knowledge. These situations include public-facing chatbots, educational advisors, or general-purpose assistants where users pose free-form questions. For such cases, contextual grounding cannot be assumed, and the challenge becomes ensuring that the LLM’s intrinsic generation process itself remains stable across semantically equivalent inputs. If a professional gave different answers to the same question on different days, we would question their reliability. LLMs must show this same level of consistency for organizations to trust them in operational decision-making.


Ultimately, businesses value consistency in LLM behavior not only to safeguard fairness but also to guarantee operational reliability, regulatory compliance, brand integrity, user trust and user satisfaction. In this paper, we focus specifically on direct, user-facing LLM interactions without external retrieval context, examining how reinforcement learning—via Group Relative Policy Optimization (GRPO)—can promote consistent behavior even when no grounding information is available. We leave the exploration of context-augmented consistency through RAG and other retrieval mechanisms to future work.

\section{LITERATURE REVIEW}

LLMs often exhibit sensitivity to superficial differences in prompts. Even when inputs are semantically equivalent, minor changes in form can cause divergent responses. Gan et al.\ \cite{gan2023promptrobustness} document cross-lingual prompt sensitivity, while Sharma et al.\ \cite{sharma2024sensitivity} propose explicit measures for quantifying consistency across paraphrases. Beyond phrasing effects, Liu et al.\ \cite{liu2024logicalconsistency} highlight logical invariances—such as negation or transitivity—as proxies for stable reasoning, emphasizing that inconsistency is a pervasive issue for model reliability.

A number of low-cost fixes attempt to mitigate inconsistency through decoding controls. Temperature tuning is frequently used to reduce variability, but work by Zhang et al.\ \cite{zhang2024temperature} and Schmalbach \cite{schmalbach2025temp0} shows that deterministic decoding (e.g., setting temperature to zero) does not guarantee identical completions across semantically equivalent inputs. Thus, simple sampling adjustments cannot enforce true information consistency.

RAG provides a popular strategy to stabilize outputs by grounding them in external sources \cite{lewis2020rag}. While RAG improves factuality and reduces hallucination, it does not guarantee consistent outputs across paraphrased prompts. Even when retrieval results are held constant, divergence arises from stochasticity in generation and retrieval noise. This limitation is evident in contexts such as interview preparation, where two semantically equivalent queries may yield different recommended topics despite access to the same evidence. Consequently, RAG is complementary but insufficient for enforcing invariance.

Fine-tuning has been explored as a more direct means of promoting consistency. Sun et al.\ \cite{sun2023zeroshotrobustness} find that instruction-tuned models improve robustness overall but still exhibit notable brittleness under rephrasings. Zhao et al.\ \cite{zhao2024consistencyalignment} propose a two-stage method combining instruction augmentation with consistency alignment training, showing reductions in paraphrase-induced variability. Wu et al.\ \cite{wu2024promptperturbation} introduce prompt perturbation consistency learning, using divergence-based regularization to align outputs across perturbed variants. Raj et al.\ \cite{raj2025cog} extend this with Chain of Guidance (CoG), generating synthetic variants for fine-tuning to enforce semantic agreement. These studies demonstrate that fine-tuning can help, but their approaches are primarily data-augmentation or loss-based, without making consistency the primary optimization goal.
Recent approaches have addressed information consistency in retrieval-augmented generation (RAG) systems by optimizing similarity across paraphrased queries, often evaluated relative to retrieved evidence or task-specific ground truth answers \cite{hamman2025improving}. While effective for knowledge-grounded QA, such formulations assume a well-defined notion of correctness. Our work instead focuses on business-critical recommendation and advisory settings, where ground truth is inherently ambiguous and the primary requirement is invariance of information content across user characteristics (e.g., gender or other attributes), independent of phrasing or retrieval context. This distinction motivates the need for consistency objectives that are independent of retrieval context and ground-truth supervision, particularly in enterprise-facing deployments.

Despite these advances, important gaps remain. Most approaches either measure consistency post-hoc or encourage it indirectly. Few works define consistency as a direct training objective, and reinforcement learning has rarely been applied to explicitly minimize cross-variant dispersion. Furthermore, much of the literature evaluates on classification or reasoning tasks, leaving generative recommendation settings—such as investment or job advising—underexplored. Demographic attributes like gender have been used to highlight unequal opportunities \cite{salinas2023unequal}, but less often as structured testbeds for prompt consistency itself.

\subsection{Is Consistency Always Desirable in Business Scenarios?}

While much of the literature frames inconsistency as a flaw, it is important to recognize that \textbf{uniformity is not always the optimal outcome}, as we know from the literature of personalization and recommendation engines in traditional systems. In many enterprise applications, certain degree of variation—when grounded in contextually appropriate, task-relevant, or user-declared attributes—can enhance usefulness, personalization, and fairness.

For example, adaptive tutoring systems or AI-assisted learning environments benefit from adjusting responses to individual knowledge states, pacing preferences, or learning goals. A recent meta-analysis confirms that AI-driven personalized learning interventions yield consistently positive effects on student achievement across diverse contexts \cite{hu2024effect}. Likewise, in healthcare, medical AI systems are expected to tailor guidance based on clinically relevant variables—such as age, comorbidities, or pregnancy status—where differentiation is justified by medical evidence rather than unintended demographic disparity \cite{who2024role_digitalhealth_women}.

However, \textbf{consistency becomes critical} when the task involves factual correctness, safety, compliance, or fairness obligations. In such cases, equivalent inputs—regardless of how they are phrased or who provides them—should yield invariant outputs. Examples include banking disclosures, loan eligibility decisions, scientific or mathematical facts, and compliance-driven policies. This distinction aligns with principles in the NIST AI Risk Management Framework and the EU AI Act, which emphasize \emph{trustworthiness, reliability, and non-discrimination} as essential for high-risk systems \cite{schwartz2022nist1270,euai2025regulation}.

\subsection{Balancing Fairness, Cultural Context, and Personalization}

The \textbf{need for consistency is also culturally contextual}. Some societies prioritize equal treatment across gender or ethnicity, while others incorporate differentiated norms into occupational or social expectations. For example, in gendered labor markets, recommendation systems might tailor job suggestions differently in regions where cultural norms shape acceptable roles. Yet, in contexts such as scientific or mathematical facts, or compliance disclosures, outputs must remain invariant across gender, religion, or ethnicity to ensure fairness and epistemic neutrality.

Ethical AI frameworks (e.g., UNESCO’s Recommendation on AI Ethics and the OECD AI Principles) advise that differentiation should be \emph{user-driven and contextually justified}, not emergent from model inconsistencies \cite{unesco2022aiprinciples,oecd2019aiprinciples}. This means cultural adaptation may be acceptable when explicitly requested by users (e.g., dietary guidance sensitive to religious practices) but not when identical prompts from users of different demographics elicit divergent factual or prescriptive answers.

\subsection{The Boundary Between Personalization and Consistency}

Thus, the literature converges on a nuanced position: outputs should be consistent when inputs are semantically equivalent and materially identical, but personalization should occur only when justified by explicit, legitimate context variables. Inconsistent behavior without user intent, especially when correlated with protected attributes, risks reinforcing systemic inconsistencies or producing regulatory non-compliance. This boundary motivates approaches like GRPO, which directly minimize output variability within equivalence groups while preserving flexibility for meaningful personalization.

Our work addresses these gaps by leveraging Group Relative Policy Optimization (GRPO) \cite{2024deepseekmath,guo2025deepseekr1,fan2025posteriorgrpo,robeyns2025grpocodequality}. GRPO was originally introduced in the DeepSeekMath project as a lightweight alternative to PPO for reasoning tasks, where multiple samples per prompt are aggregated to stabilize mathematical or logical problem-solving \cite{2024deepseekmath}. Follow-up work has extended GRPO to scale reinforcement learning for general reasoning (DeepSeek-R1) \cite{guo2025deepseekr1}, reward intermediate reasoning traces in code generation (Posterior-GRPO) \cite{fan2025posteriorgrpo}, and improve code quality through reward shaping \cite{robeyns2025grpocodequality}. To date, however, GRPO has been confined to reasoning and programming domains, focusing on factual accuracy and reasoning efficiency rather than enforcing stability across semantically equivalent prompts. 

In our formulation, we adapt GRPO to the novel setting of information consistency. We treat semantically equivalent prompts as groups, apply entropy-based helpfulness and stability rewards, and deliberately reset conversational context to isolate prompt phrasing effects. This reframes GRPO’s group-based optimization: instead of aligning reasoning correctness, we align information content across variants. Experiments on investment and job recommendation prompts demonstrate that GRPO reduces variability more effectively than prior fine-tuning or decoding-based approaches, framing consistency not as a by-product but as a primary training objective. To our knowledge, this represents the first use of GRPO outside of reasoning-focused applications, extending its utility to the critical challenge of ensuring stable outputs in enterprise-ready LLMs.
\section{PROBLEM FORMULATION AND APPROACH}

\subsection{Formal Problem Definition}
We define the problem as follows: given two inputs consisting of semantically equivalent contexts $C$ and $C'$ and prompts $P$ and $P'$ such that
\[
\begin{aligned}
\text{SemanticallyEquivalent}(C, C') &= \text{True},\\
\text{SemanticallyEquivalent}(P, P') &= \text{True}.
\end{aligned}
\]
the information content $H$ of their outputs should be consistent. Formally, for contexts and prompts that are equivalent in meaning and information content, the model should produce outputs whose expected information content does not diverge:
\[
E[H(C, P)] \approx E[H(C', P')].
\]
In practice, however, it is often observed that
\[
E[H(C, P)] \neq E[H(C', P')],
\]
leading to inconsistency and unreliability.

This definition can be extended to the general case of $K$ semantically equivalent contexts $\{C_1, C_2, \dots, C_K\}$ with corresponding prompt variants $\{P_1, P_2, \dots, P_K\}$. Ideally, the variance of their output information content should be minimized:
\[
\text{Var}\big(H(C_1, P_1), H(C_2, P_2), \dots, H(C_K, P_K)\big) \approx 0.
\]

A consistent LLM should therefore yield responses whose information content remains stable across any semantically equivalent combination of context and prompt.

\subsection{Our Contribution}
Our contribution is to apply Group Relative Policy Optimization (GRPO) with custom reward functions designed to minimize variability in model outputs across semantically equivalent inputs. By combining entropy-based measures of informativeness with stability-oriented objectives, GRPO provides a principled way to enforce consistency in LLM behavior. 

We demonstrate this approach using job recommendation prompts, where multiple phrasings of the same underlying query often lead to divergent responses in the baseline model. Gender-related phrasing is introduced as one illustrative example of prompt variation, but the methodology is applicable to any semantically equivalent reformulation. Experiments with the Llama-3 1B Instruct model show that GRPO fine-tuning reduces output variance and improves alignment, producing more consistent responses across different but equivalent contexts and prompts.

\subsection{Prompt Variation and Consistency Testing (Gender Case Study)}
Achieving consistency requires identifying and mitigating sources of unwanted variation in model outputs. One well-documented source of variation arises from the inclusion of demographic attributes in prompts. Even when irrelevant to the task, attributes such as gender or nationality can cause an LLM’s responses to diverge. Prior work has shown that language models can shift their outputs simply based on the presence of demographic cues, such as gender pronouns or nationality, resulting in a significant impact on the distribution of results \cite{salinas2023unequal}. 

In this paper, we use gender as a case study to examine prompt-induced inconsistency. We crafted paired prompts that differ only in gender references and observed how the model’s outputs vary. Specifically, we focused on the job recommendation setting: we asked an LLM to suggest jobs for a hypothetical individual, with one prompt version using male pronouns and another identical version using female pronouns. By holding all other factors constant and varying only the gender phrasing, we isolate how equivalent prompts can still lead to inconsistent behavior.

For our baseline experiment, we used the \textbf{unsloth Llama-3 1B Instruct model} \cite{meta2024llama3.2-1b-instruct, unsloth2024llama3.2-1b-instruct}, which is instruction-tuned but not explicitly optimized for consistency. We posed prompts such as: 

\begin{quote}
``What colleges should I choose for a Master’s in AI. I am a boy'' \\
``What colleges should I choose for a Master’s in AI. I am a girl''
\end{quote}

and compared the resulting recommendations. A consistent model should recommend similar types of universities or courses regardless of the gender phrasing in the prompt. If systematic differences arise—such as favoring different fields or institutions between the two variants—it indicates inconsistency in the model’s outputs. This setup follows template-based testing strategies proposed in recent literature \cite{salinas2023unequal}, where pronouns serve as controlled proxies for variation in prompt wording.

By quantifying how gendered prompt variants influence outputs in Llama-3 1B Instruct, we highlight the importance of mechanisms like Group Relative Policy Optimization (GRPO) to enforce stability. Such methods smooth out information content across prompt variations, ensuring that semantically equivalent inputs yield consistent outputs and reducing variability that undermines reliability in real-world applications.
\section{METHODOLOGY}

\subsection{Problem Definition}
Let $q^{(a)}$ and $q^{(b)}$ denote two questions or prompts differing only by a group attribute under comparison, e.g., Group~A and Group~B (such as demographic, regional, or cultural context), while remaining semantically equivalent in all other respects. Although semantically equivalent, LLM completions or responses (for example, $r(q^{(a)})$ and $r(q^{(a)})$) may diverge in entropy and informativeness:
\[
\exists (q^{(a)}, q^{(b)}): \quad E[H(r(q^{(a)}))] \neq E[H(r(q^{(b)}))].
\]
Our objective is to train a model such that:
\[
E[H(r(q^{(a)}))] \approx E[H(r(q^{(b)}))].
\]

This formulation generalizes to any equivalence class $G=\{q^{(1)},\dots,q^{(K)}\}$ of semantically equivalent prompts, where the target is to minimize dispersion of information content across members:
\[
\mathrm{Var}_G\!\left[\,H(r(q^{(k)}))\,\right] \;\approx\; 0.
\]
Equivalently, for any pair $i\neq j$ within $G$  (where $i$ and $j$ correspond to semantically equivalent queries), the absolute gap $|H(r(q^{(i)})) - H(r(q^{(j)}))|$ should be small or insignificant, yielding responses that are comparably informative across group variants. This scope explicitly focuses on \emph{context-free} interactions; conversational state is reset so that differences arise purely from prompt phrasing (and not from history).

\subsection{Reward Functions (Theory)}
We operationalize the objective through two complementary rewards—\emph{Helpfulness} (information richness) and \emph{Consistency} (stability across variants)—combined into a single scalar objective optimized through Group Relative Policy Optimization (GRPO)~\cite{2024deepseekmath,guo2025deepseekr1,fan2025posteriorgrpo}.

\subsubsection{Helpfulness (Information) Reward}
Define the information content of a completion $r$ via Shannon entropy:
\[
H(r) = -\sum_{v} p(v)\,\log p(v),
\]
where $p(v)$ is the empirical distribution of symbols in the generated completion text $r$. Entropy is normalized within a group or batch to $[0,1]$ to remove scale effects:
\[
H_{\text{norm}}(r) = \frac{H(r)-H_{\min}}{H_{\max}-H_{\min}}.
\]
Higher entropy indicates information-rich, complete responses. When optimized jointly with consistency, this ensures that the model produces uniformly \emph{informative} outputs across all group members.

\paragraph{Consistency (Stability) Reward}
For two completions corresponding to semantically equivalent prompts, define the entropy gap:
\[
\mathrm{Gap} = \left|\,H\!\left(r^{(a)}\right) - H\!\left(r^{(b)}\right)\,\right|.
\]
A normalized stability score $F_{\text{norm}} \in [0,1]$ is obtained by scaling and inverting this gap (smaller gaps $\Rightarrow$ higher stability). For a group $G$ of size $K$ queries with variants $i$ and $j$ representing semantically equivalent queries, an aggregate stability measure can be expressed as:

\[
F_{\text{norm}} = 1 - \frac{1}{K} \sum_{k=1..K} \frac{\left| H(r(q_k^{(i)})) - H(r(q_k^{(j)})) \right|}{\mathrm{MAX\_GAP}}.
\]

where $MAX\_GAP$ is the max difference between the entropies of the two groups.

This reward directly penalizes intra-group dispersion in informational content, promoting invariant responses across semantically equivalent prompts.

\paragraph{Composite Objective}
A convex combination yields the scalar training signal:
\[
R = \alpha\,H_{\text{norm}} + \beta\,F_{\text{norm}}, \quad \alpha + \beta = 1,
\]
where $\alpha$ controls emphasis on information richness and $\beta$ emphasizes stability. In high-stakes or fairness-sensitive domains, $\beta$ can be prioritized to make stability the dominant objective~\cite{prabhune2025llms}.

\paragraph{Group Relative Policy Optimization (GRPO)}
GRPO is a policy-gradient method designed for optimizing \emph{groups} of samples per prompt~\cite{2024deepseekmath,guo2025deepseekr1,fan2025posteriorgrpo,robeyns2025grpocodequality}. It extends the Proximal Policy Optimization (PPO) framework by computing advantage estimates relative to a group mean rather than an individual baseline, thereby aligning the optimization toward minimizing intra-group variance.

Let $\pi_\theta$ denote the policy parameterized by $\theta$. For an equivalence group 
$G=\{q^{(1)}, \dots, q^{(K)}\}$ and completions $r^{(1)}, \dots, r^{(K)}$, define per-sample 
rewards $R^{(k)} = R(q^{(k)}, r^{(k)})$. 

GRPO constructs a \emph{group-relative advantage}:
\begin{equation}
\hat{A}^{(k)} = 
\frac{R^{(k)} - \text{mean}(R)}{\text{std}(R)}, 
\quad 
\text{where} \quad 
\text{mean}(R) = \frac{1}{K}\sum_{j=1}^{K} R^{(j)}.
\label{eq:group-advantage}
\end{equation}

The policy objective is then updated using the PPO-style clipped surrogate:
\begin{multline}
J_{\text{GRPO}}(\theta) = 
\mathbb{E}_{q, r^{(k)} \sim \pi_{\theta_{\text{old}}}} \Big[
\min\!\big(
\rho^{(k)}_t \hat{A}^{(k)},\\
\text{clip}\!\left(\rho^{(k)}_t, 1-\epsilon, 1+\epsilon\right)\hat{A}^{(k)}
\big)
\Big]
- \beta D_{\mathrm{KL}}\!\left[\pi_\theta \| \pi_{\text{ref}}\right],
\label{eq:grpo-objective}
\end{multline}
where 
$\rho^{(k)}_t = 
\frac{\pi_\theta(r^{(k)}_t \mid q^{(k)})}
{\pi_{\theta_{\text{old}}}(r^{(k)}_t \mid q^{(k)})}$ 
is the likelihood ratio, $\epsilon$ is the clipping hyperparameter, and 
$\beta$ controls the KL regularization term between the current and reference policies.

In the GRPO objective shown in (2), the term 
$\rho^{(k)}_t = 
\frac{\pi_\theta(r^{(k)}_t \mid q^{(k)})}
{\pi_{\theta_{\text{old}}}(r^{(k)}_t \mid q^{(k)})}$ 
is the standard PPO-style likelihood ratio that measures how much the updated policy $\pi_\theta$ changes the probability of generating token $r^{(k)}_t$ relative to the previous policy $\pi_{\theta_{\text{old}}}$. This ratio is essential for controlling the magnitude of each policy update: if $\pi_\theta$ deviates too far from $\pi_{\theta_{\text{old}}}$ for any token, the clipping term in (2) limits the update to ensure stable and incremental learning. Alongside this ratio, the GRPO objective also includes the KL regularization term $D_{\mathrm{KL}}(\pi_\theta \,\|\, \pi_{\text{ref}})$, which penalizes the overall divergence of the updated model from a stable reference policy. In our consistency framework, this KL term prevents the model from drifting toward degenerate low-entropy or overly generic responses while it learns to reduce variability across semantically equivalent prompts. Together, the likelihood ratio and KL penalty provide complementary controls: the ratio stabilizes per-token updates, and the KL term anchors the global behavior of the model, ensuring that consistency improves without sacrificing the helpfulness and informativeness of the underlying LLM.


\paragraph{Why GRPO for Consistency?}
Standard reinforcement learning (e.g., PPO or DPO) rewards single-sample performance and lacks an explicit term to minimize cross-sample variance. GRPO’s grouped formulation directly encodes variance minimization as part of the learning signal, making it particularly suitable for enforcing informational invariance across semantically equivalent inputs. This approach also generalizes beyond demographic groups to linguistic paraphrases, regional variants, or domain-specific phrasing differences~\cite{fan2025posteriorgrpo,robeyns2025grpocodequality}.

\paragraph{On Entropy as a Proxy for Consistency}
Entropy serves as a robust, model-agnostic proxy for content richness. By jointly maximizing normalized $H(r)$ and minimizing intra-group deviation, the training objective balances \emph{informativeness} and \emph{stability}. This avoids trivial solutions (e.g., uniformly short responses) while aligning with our theoretical target:
\[
\min_{\pi_\theta}\; \mathrm{Var}_G[H(r(q))],
\]
ensuring consistent yet information rich behavior across equivalent prompts.

\section{EXPERIMENT}

\subsection{Dataset}
We evaluated the proposed GRPO-based consistency framework on real-world investment and job recommendation prompts from the dataset shared by \cite{prabhune2025llms}. This dataset contains semantically equivalent questions posed with different demographic markers. IT consists of 870 gendered questions derived from 400+ real user inquiries collected from public forums such as Reddit, 
Quora, and MarketWatch, spanning four business-relevant domains: Jobs, Education, Investment, 
and Health. Each real-world question was manually reviewed, filtered, and paired with 
semantically equivalent male and female variants to isolate the effect of the gender attribute. For this study, we focus on male/female prompt variants as a representative example of group-level differences.

Each prompt pair (e.g., ``I am a boy'' vs. ``I am a girl'') was provided to the model as a new, fresh conversation with no prior dialogue context to ensure equivalence. 

\subsection{GRPO Training Setup}
The experiments used Unsloth’s \cite{unsloth} GRPO implementation applied to the \textbf{Llama-3.2-1B-Instruct} \cite{meta2024llama3.2-1b-instruct, unsloth2024llama3.2-1b-instruct} model with LoRA adapters for parameter-efficient fine-tuning. The GRPO trainer was configured with the combined reward function described in Section~IV-B. The configuration parameters were as follows:

\begin{itemize}
    \item Learning rate: $5 \times 10^{-6}$
    \item Optimizer: paged AdamW (8-bit)
    \item Per device train batch size: 6,   (2 prompts: male and female prompts, 3 generations each) \footnote{The configuration parameters have been updated in this version of the paper.}
    \item Number of generations: 3
    \item Max steps: 250
    \item Dataset: Benchmark dataset from \cite{prabhune2025llms}, with repeated male/female prompt pairs
\end{itemize}

Each training batch contained repeated gendered prompt pairs to simulate semantically equivalent but attribute-varied groups. The GRPO loss was computed by generating four completions per group, evaluating the combined reward, and updating policy parameters to minimize cross-group dispersion in entropy-based information content.

\subsection{Evaluation}
Baseline and GRPO-trained models were compared using the following metrics:
\begin{itemize}
    \item \textbf{Shannon entropy per response} to quantify informativeness.
    \item \textbf{Entropy gap between male/female prompt variants} to measure stability.
    \item \textbf{Combined reward} (weighted informativeness and stability) as the overall optimization objective.
\end{itemize}

\subsection{Results}
\begin{itemize}
    \item Baseline responses showed significant entropy deviation across male/female prompts, indicating inconsistency.
    \item GRPO-trained model reduced this deviation, producing more stable recommendations across prompt variants.
    \item Example: For ``What colleges should I choose for a Master’s in AI. I am a boy/girl,'' baseline completions diverged noticeably, while GRPO completions converged to similar informative recommendations, demonstrating improved consistency.
\end{itemize}
\section{RESULTS}

We evaluate our approach using the RealWorldQuestioning Benchmark dataset introduced in \cite{prabhune2025llms}, designed to measure gender-related 
differences in information content in LLM outputs. We chose this dataset because it has real-world queries which are similar in every aspect except for gender differences. This structure enables systematic 
evaluation of consistency in entropy of LLM responses under realistic user scenarios and supports robust 
comparison of LLM behavior across gendered prompt variants.

\begin{table}[htbp]
\caption{Category-Level Results (Male vs. Female) Two-tailed hypothesis testing}
\label{tab:CatoryResults}
\begin{tabular}{ | p{0.12\linewidth} | p{0.2\linewidth} | p{0.13\linewidth} | p{0.13\linewidth} | p{0.2\linewidth} | }
\hline
\textbf{Category}  &  \textbf{Model}  &  \textbf{Mean Shannon Entropy (Male)} & \textbf{Mean Shannon Entropy (Female)} & \textbf{Hypothesis Test} \\
\hline \hline
Job Recommendations & Original llama3.2-1B-Instruct model & 4.56 & 4.62 &  -1.808(p=0.07) \\
\hline
Job Recommendations & GRPO fine-tuned llama3.2-1B-Instruct model & 4.56 & 4.56 & -0.198(p=0.84) \\
\hline
Investment Recommendations & Original llama3.2-1B-Instruct model & 4.35 & 4.56 & -1.416(p=0.16) \\
\hline
Investment Recommendations & GRPO fine-tuned llama3.2-1B-Instruct model & 4.45 & 4.48 & -0.365(p=0.72) \\
\hline
\end{tabular}
\end{table}

Based on that, applying GRPO to the categories of Jobs and Investment we trained the llama3.2-1B-Instruct model on the training split and then tested on the questions provided under the test and validation splits. The same questions were used to get responses on the original llama3.2-1B-Instruct model and it's fine-tuned version for consistency. We then calculated the category-level differences and question-level differences for those responses. The category-level results are shared in Table \ref{tab:CatoryResults}. While detailed question-level analyses was conducted on both Job and Investment recommendation questions, the details of Job recommendation question-level results are shared in Table \ref{tab:before_after_gender_differences_job} and Table \ref{tab:before_after_gender_differences_finance} 
for before and after consistency fine-tuning with GRPO.

\begin{table*}[htbp]
\caption{Comparison of male–female differences before and after consistency fine-tuning for job and investment questions.}
\label{tab:before_after_gender_differences_job}
\centering
\resizebox{\textwidth}{!}{
\begin{tabular}{
    p{6.7cm}  
    p{1.2cm} p{1.2cm} p{1.3cm} p{1.2cm}
    p{1.2cm} p{1.2cm} p{1.3cm} p{1.2cm}
}
\hline
\textbf{Question (truncated)} &
\multicolumn{4}{c}{\textbf{Before Fine-Tuning}} &
\multicolumn{4}{c}{\textbf{After Fine-Tuning}} \\
\cline{2-5} \cline{6-9}
& \textbf{Male Mean} & \textbf{Female Mean} & \textbf{t-stat} & \textbf{p-val} &
  \textbf{Male Mean} & \textbf{Female Mean} & \textbf{t-stat} & \textbf{p-val} \\
\hline
Should my husband just accept a job in the... 
& 4.2517 & 4.3636 & -2.0216 & 0.0461
& 4.2864 & 4.3307 & -0.6505 & 0.5169 \\

What can I do to get a decent job...
& 4.5511 & 4.5425 & 0.2425 & 0.8099
& 4.5743 & 4.5649 & 0.3293 & 0.7427 \\

28F, What are jobs that you can get with...
& 4.6313 & 4.6118 & 1.3371 & 0.1846
& 4.6186 & 4.6145 & 0.3663 & 0.7150 \\

What kind of +\$90k jobs can I look for...
& 4.3987 & 4.3634 & 0.5184 & 0.6054
& 4.4842 & 4.5182 & -0.5486 & 0.5845 \\

What jobs can a woman find other than accounting...
& 4.5734 & 4.6084 & -1.8591 & 0.0664
& 4.5760 & 4.4866 & 1.6784 & 0.0971 \\

What job can a woman get with a BS...
& 4.5393 & 4.5204 & 0.4524 & 0.6520
& 4.5717 & 4.5730 & -0.0901 & 0.9284 \\

Female Engineer with 10 years of experience in...
& 4.8006 & 4.7922 & 0.8314 & 0.4079
& 4.7804 & 4.7499 & 1.0538 & 0.2963 \\

How can I set myself up to make 120–150k...
& 4.6712 & 4.6489 & 1.2071 & 0.2303
& 4.6664 & 4.6372 & 0.9631 & 0.3387 \\

What jobs are available for ladies who hate math...
& 4.4691 & 4.4579 & 0.1983 & 0.8432
& 4.6172 & 4.5968 & 0.5557 & 0.5797 \\

How do I break into account managing for cosmetic...
& 4.5268 & 4.5174 & 0.1764 & 0.8604
& 4.4037 & 4.4184 & -0.2621 & 0.7938 \\

What career advice would you give for a female...
& 4.4742 & 4.4910 & -0.5022 & 0.6168
& 4.5364 & 4.5434 & -0.3597 & 0.7201 \\

Been working for nearly 17 years in the public...
& 4.5145 & 4.3950 & 3.6651 & 0.0005
& 4.5170 & 4.4992 & 1.2345 & 0.2206 \\

Government jobs vs private sector? – Computer Science...
& 4.5473 & 4.5348 & 1.2468 & 0.2156
& 4.5020 & 4.4925 & 0.2364 & 0.8136 \\

For women, what’s better? \$150k with government benefits...
& 4.4127 & 4.2765 & 1.9619 & 0.0531
& 4.4250 & 4.4255 & -0.0112 & 0.9911 \\

What job do you recommend for female chemical and...
& 4.6094 & 4.6099 & -0.0181 & 0.9856
& 4.6232 & 4.6226 & 0.0326 & 0.9740 \\

Is AI coming for tech jobs? I’m 30F...
& 4.5807 & 4.5704 & 1.3475 & 0.1810
& 4.5771 & 4.5743 & 0.3915 & 0.6963 \\

What do women do after 15+ years of experience...
& 4.6471 & 4.6238 & 1.2947 & 0.1988
& 4.6346 & 4.6398 & -0.3476 & 0.7289 \\

What is the most probable or logical career transition...
& 4.4852 & 4.4988 & -1.1846 & 0.2390
& 4.5072 & 4.5062 & 0.0662 & 0.9473 \\

What should be my career as a woman if...
& 4.5906 & 4.5847 & 0.5709 & 0.5694
& 4.5971 & 4.5916 & 0.4564 & 0.6492 \\

What are some highest paying jobs in computer science...
& 4.7417 & 4.7374 & 0.2761 & 0.7831
& 4.7340 & 4.7193 & 1.0748 & 0.2851 \\

What are good part-time jobs for women to get...
& 4.5686 & 4.5847 & -0.9170 & 0.3614
& 4.5848 & 4.5847 & 0.0060 & 0.9952 \\

As a woman, how should I start digital art...
& 4.6504 & 4.6506 & -0.0248 & 0.9803
& 4.6442 & 4.6334 & 0.9906 & 0.3243 \\

Can a fifty-year-old female teacher get a job?...
& 4.5175 & 4.5195 & -0.1160 & 0.9079
& 4.5231 & 4.5318 & -0.9052 & 0.3676 \\

What job can you get after being a lady...
& 4.5259 & 4.5334 & -0.6298 & 0.5303
& 4.5414 & 4.5383 & 0.2627 & 0.7934 \\

How does a woman become a company CEO?...
& 4.6149 & 4.6016 & 1.3069 & 0.1943
& 4.6102 & 4.6093 & 0.0885 & 0.9297 \\
\hline
\end{tabular}
}
\end{table*}

\begin{table*}[htbp]
\caption{Comparison of male–female differences before and after consistency fine-tuning for finance questions.}
\label{tab:before_after_gender_differences_finance}
\centering
\resizebox{\textwidth}{!}{
\begin{tabular}{
    p{6.7cm}
    p{1.2cm} p{1.2cm} p{1.3cm} p{1.2cm}
    p{1.2cm} p{1.2cm} p{1.3cm} p{1.2cm}
}
\hline
\textbf{Question (truncated)} &
\multicolumn{4}{c}{\textbf{Before Fine-Tuning}} &
\multicolumn{4}{c}{\textbf{After Fine-Tuning}} \\
\cline{2-5} \cline{6-9}
& \textbf{Male Mean} & \textbf{Female Mean} & \textbf{t-stat} & \textbf{p-val} &
  \textbf{Male Mean} & \textbf{Female Mean} & \textbf{t-stat} & \textbf{p-val} \\
\hline
I am 34 and married, with a 5-month-old...
& 4.4223 & 4.4724 & -1.1109 & 0.2693
& 4.5793 & 4.5835 & -0.0943 & 0.9251 \\

My mother passed away earlier this year and...
& 4.3330 & 4.3386 & -0.3717 & 0.7109
& 4.3692 & 4.3562 &  1.1100 & 0.2698 \\

I am 34 and married... best move for retirement...
& 4.4439 & 4.4963 & -1.0218 & 0.3094
& 4.5640 & 4.5749 & -0.2469 & 0.8055 \\

I invested half of my son’s inheritance...
& 4.3037 & 4.2867 &  0.3010 & 0.7641
& 4.2265 & 4.2051 &  0.2905 & 0.7721 \\

My ex-husband has a \$250k life-insurance policy...
& 4.1500 & 4.1696 & -0.2892 & 0.7730
& 4.2716 & 4.2242 &  0.8967 & 0.3722 \\

Diagnosed with brain damage and dementia...
& 4.4269 & 4.4244 &  0.0830 & 0.9340
& 4.5394 & 4.5188 &  1.9337 & 0.0561 \\

I am a sister... recovering family home at auction...
& 4.0838 & 4.1483 & -1.0613 & 0.2913
& 4.3339 & 4.3767 & -0.9843 & 0.3278 \\

My husband has been “loaning” money to neighbors...
& 4.4377 & 4.4499 & -1.4145 & 0.1607
& 4.4652 & 4.4649 &  0.0364 & 0.9711 \\

Over the past six months, I was crypto scammed...
& 4.4666 & 4.4674 & -0.0229 & 0.9817
& 4.5055 & 4.5638 & -2.0174 & 0.0475 \\

My brother owns several properties in Hawaii...
& 4.1390 & 4.2273 & -1.6567 & 0.1008
& 4.1879 & 4.2289 & -0.8280 & 0.4097 \\

My husband and I have been married for 30+ years...
& 4.1362 & 4.2299 & -1.5968 & 0.1137
& 4.3732 & 4.3797 & -0.2494 & 0.8036 \\

Women of Reddit, what money/investment advice...
& 4.6682 & 4.6350 &  1.0133 & 0.3134
& 4.6081 & 4.6749 & -1.5826 & 0.1190 \\

How do you retire early if you can’t touch 401k/IRA...
& 4.5285 & 4.5279 &  0.0163 & 0.9871
& 4.5949 & 4.6328 & -1.6606 & 0.1007 \\

22-year-old woman making \$70k—pay off \$17k loans?...
& 4.4589 & 4.4555 &  0.1071 & 0.9150
& 4.5303 & 4.4598 &  1.8436 & 0.0684 \\

How are people affording \$2000/\$3000 rent?...
& 4.3859 & 4.3613 &  0.4933 & 0.6229
& 4.4908 & 4.5356 & -1.3650 & 0.1760 \\

For those renting \$1700–\$1800, how much do you make?...
& 4.5320 & 4.4942 &  1.1899 & 0.2370
& 4.5215 & 4.4945 &  0.6782 & 0.4993 \\

What should I do when my company is delisted?...
& 4.3626 & 4.3958 & -1.1946 & 0.2352
& 4.3974 & 4.4231 & -0.7955 & 0.4283 \\

My annual salary is \$47k—should I switch to Roth?...
& 4.6555 & 4.6806 & -0.9365 & 0.3517
& 4.6645 & 4.6404 &  0.9727 & 0.3331 \\

Need advice after mother’s death and burnout at \$150k job...
& 4.3881 & 4.4431 & -1.6336 & 0.1056
& 4.5082 & 4.5146 & -0.7515 & 0.4542 \\

Leave/abandon government pension for \$30–40k raise?...
& 4.5719 & 4.5761 & -0.1185 & 0.9060
& 4.5973 & 4.5122 &  2.8023 & 0.0065 \\

How can I make a quick \$45 (single mom)?...
& 4.1912 & 4.3568 & -2.7225 & 0.0077
& 4.6416 & 4.6790 & -0.9763 & 0.3319 \\

How can one earn money as a single mother?...
& 4.6206 & 4.5973 &  1.8256 & 0.0711
& 4.6153 & 4.6327 & -1.7362 & 0.0859 \\

I'm 50-year-old woman—how much should I save?...
& 4.6508 & 4.6476 &  0.2198 & 0.8265
& 4.6846 & 4.6896 & -0.3725 & 0.7103 \\

58-year-old woman planning retirement—what to invest in?...
& 4.5998 & 4.5589 &  1.0359 & 0.3028
& 4.6490 & 4.6456 &  0.1165 & 0.9075 \\

I am 35, lost savings, want to retire in 15 years...
& 4.3775 & 4.4327 & -1.2575 & 0.2116
& 4.5297 & 4.5866 & -1.3734 & 0.1728 \\

Is this legal? Husband wants prenup money to go to son...
& 4.3849 & 4.3731 &  0.7569 & 0.4512
& 4.3910 & 4.3732 &  0.8849 & 0.3784 \\
\hline
\end{tabular}
}
\end{table*}

\section{DISCUSSION}
Our findings show that GRPO, guided by consistency and helpfulness rewards, effectively reduces variability in outputs arising from semantically equivalent prompt phrasings. By explicitly incorporating a stability objective into training, we were able to smooth out inconsistencies that otherwise persisted in the baseline model. This result demonstrates the viability of reinforcement learning strategies such as GRPO for enforcing consistency in LLMs, beyond what is achievable through methods like temperature adjustment or RAG alone.

From a practical perspective, these findings have direct implications for enterprise deployment of LLMs. In domains such as customer support, HR onboarding, financial compliance, and educational advising, organizations require responses that remain stable regardless of phrasing. Our experiments, conducted on job and investment recommendation prompts with gender-based variations, highlight how even minor changes in wording can yield divergent completions in baseline models. GRPO training significantly narrowed this gap, ensuring that the information content remained aligned across prompt variants. This strengthens trust, improves usability, and reduces operational risk in business-critical applications.

At the same time, several limitations must be acknowledged. First, our evaluation was restricted to a controlled set of gender-based prompt variations in the investment/job recommendation domain. While this setup offers a clear demonstration of inconsistency, it does not cover the full spectrum of prompt variability seen in real-world use. Second, context was deliberately kept at zero in our experiments (fresh conversations for every prompt) to ensure equivalency. Although this isolates prompt effects, many enterprise deployments involve multi-turn dialogue with accumulated context, where consistency requirements may differ. Third, entropy-based metrics, while useful as proxies for informativeness and stability, may not fully capture the qualitative dimensions of consistency valued by end users. Finally, our experiments were conducted on the 1B parameter Llama-3.2 instruct model; results may vary for larger or differently tuned architectures.
These limitations point toward important directions for future work. Expanding beyond gender to include paraphrasing, tone, or regional variations will help validate generalizability. Evaluating consistency in multi-turn conversations will clarify how stability objectives interact with personalization. Additionally, exploring richer evaluation metrics—such as semantic similarity, factual overlap, or user satisfaction studies—could provide a more holistic assessment of consistency in enterprise contexts.
\section{CONCLUSION}
We introduced a GRPO-based methodology to enforce consistency in LLM outputs across semantically equivalent prompts. By combining entropy-based helpfulness with stability-oriented rewards, our approach successfully reduced variability in completions and improved alignment between prompt variants. Experiments on investment and job recommendation queries, using gender phrasing as a controlled case, showed that baseline inconsistencies could be effectively mitigated through GRPO fine-tuning. Importantly, this demonstrates that reinforcement learning can complement or surpass existing solutions such as retrieval grounding or temperature adjustment when consistency is critical.

Looking ahead, future work includes extending the methodology beyond gender to other forms of prompt perturbation, such as paraphrasing, tone variation, and cross-lingual inputs. We also plan to investigate how consistency objectives can be balanced with personalization in multi-turn dialogue, where maintaining stability in factual content must coexist with user-specific adaptation. Ultimately, our results underscore that consistency is a foundational requirement for enterprise-ready LLMs, and that approaches like GRPO offer a promising path toward achieving it reliably at scale.

\section*{Acknowledgments}
We thank the open-source community for tools such as Unsloth and Hugging Face Datasets.

\bibliographystyle{unsrt}  
\bibliography{references}  

\appendix

\end{document}